%% file: Template.tex
\documentclass{article}
\usepackage[utf8]{inputenc}
\usepackage{spconf,amsmath,amssymb,graphicx}
\usepackage{times}
\usepackage{booktabs}
\usepackage{multirow}
\usepackage{array}
\usepackage{tabularx}
\usepackage{url}
\usepackage{cite}
\usepackage{algorithm}
\usepackage{algorithmic}
\usepackage{framed}
\usepackage[pagebackref=false,breaklinks=true,letterpaper=true,colorlinks,bookmarks=false]{hyperref}

\newcommand{\mesh}{\textsc{MESH}}
\newcommand{\method}{\mesh}
\newcommand{\meshb}{\textsc{MESH-B}}
\newcommand{\meshfull}{\textsc{MESH-Full}}
\newcommand{\meshbfull}{\textsc{MESH-B-Full}}

\newcommand{\smalltab}{\small}

\title{MESH: Memory-Efficient Sinkhorn Optimization for Mixture-of-Experts Training}

\name{Masato Fujitake}
\address{FA Research, Fast Accounting Co., Ltd.\\
{\tt\small fujitake@fastaccounting.co.jp}}

\begin{document}
\maketitle

\begin{abstract}
\input{files/abstract}
\end{abstract}


\section{Introduction}
\label{sec:intro}
\input{files/introduction}

\section{Related Work}
\label{sec:related}
\input{files/related}

\section{Background and Diagnosis}
\label{sec:background}
\input{files/background}

\section{MESH}
\label{sec:method}
\input{files/method}

\section{Experiments}
\label{sec:experiments}
\input{files/experiments}

\section{Analysis and Limitations}
\label{sec:analysis}
\input{files/analysis}

\section{Conclusion}
\label{sec:conclusion}
\input{files/conclusion}

\clearpage
\onecolumn
\appendix
\section{Additional Experimental Results}
\label{app:results}
\input{files/appendix_results}

\clearpage
\section{Pseudocode and Implementation Details}
\label{app:algo}
\input{files/appendix_algorithm}

\end{document}

%% file: files/abstract.tex
Memory-efficient matrix optimizers such as Sinkhorn gradient descent remove most AdamW optimizer state for dense Transformer matrices, but direct application to Mixture-of-Experts (MoE) training is unreliable.  
We study this failure in a controlled 110M-parameter nanowhale DeepSeek-style MoE pretraining setting.  
A SAGE/Sinkhorn hybrid reduces optimizer state from 0.883GB to 0.331GB but degrades evaluation loss to 3.8265, far above the AdamW baselines observed in the same setup (3.58--3.64 across the seeds we study).  
We show that routed MoE expert matrices are the dominant failure point: their gradients are conditional, temporally varying, and poorly served by stateless Sinkhorn normalization.  
We propose \mesh{}, a hidden-momentum Sinkhorn update for MoE experts.  
\mesh{} restores a temporal first-moment signal through the gradient-buffer lifecycle, without storing the expert first moment as optimizer state.  
\meshb{} is an optional block-preconditioned variant that adds a coarse neuron/block inverse-RMS multiplier.  
Across ablations, temporal smoothing before matrix normalization is the primary causal ingredient; block/neuron preconditioning can improve the memory-quality frontier, but is not established as universally necessary.  
In two additional seeds, \mesh{} and \meshb{} reduce optimizer-state memory by 62.5\% and peak PyTorch CUDA allocation by about 12.6\% relative to AdamW, with a modest evaluation-loss gap.  
Full-state diagnostic variants recover AdamW-like performance in ablations, supporting the conclusion that MoE experts need temporal smoothing, but not necessarily full coordinate-wise AdamW state.

%% file: files/introduction.tex
AdamW remains the default optimizer for language-model pretraining because it combines temporal first-moment smoothing with coordinate-wise adaptive scaling~\cite{kingma2015adam,loshchilov2019decoupled}.  
Its memory cost, however, is substantial: AdamW stores two full state tensors per trainable parameter.  
This motivates optimizers that reduce or remove moment state.  
Sinkhorn-style matrix optimizers normalize gradient matrices by alternating row and column operations, yielding stateless updates for dense two-dimensional weights~\cite{scetbon2025gradient}.  
SAGE-style hybrids further recognize that embeddings are difficult for purely stateless methods and assign light-state updates to vocabulary tensors while leaving dense matrices to Sinkhorn~\cite{lee2026sage}.

These role assignments were developed mainly for dense Transformer settings.  
We ask whether they transfer to Mixture-of-Experts (MoE) language models.  
The answer is negative: a routed expert matrix is two-dimensional, but its gradient is not the same object as a dense MLP gradient.  
Each expert receives a router-selected token subset, weighted by learned gate scores, and this conditional distribution changes across training.  
In a nanowhale 110M DeepSeek-style MoE model, a direct SAGE/Sinkhorn hybrid lowers optimizer state to 0.331GB but reaches only 3.8265 evaluation loss, far worse than AdamW.  
Applying SAGE to experts is worse still, and reducing Sinkhorn scale, fusing expert tensors, or using route-confidence damping does not close the gap.

The key question is therefore not whether Sinkhorn can be memory-efficient, but what it lacks for routed MoE experts.  
Our ablations indicate that the primary missing ingredient is temporal smoothing before matrix normalization.  
Stateless Sinkhorn computes $\mathrm{Sinkhorn}(G_t)$ from the instantaneous expert gradient.  In a routed expert, $G_t$ is a high-variance, conditional sample of an expert-specific token distribution.  Because Sinkhorn is nonlinear, $\mathrm{Sinkhorn}(\mathbb{E}[G_t])$ and $\mathbb{E}[\mathrm{Sinkhorn}(G_t)]$ can behave differently; the latter amplifies step-to-step routing noise.  
The optimizer failure is an ordering problem: normalize-then-average behaves differently from average-then-normalize.  
\mesh{} converts the update to $\mathrm{Sinkhorn}(H_t)$, where $H_t$ averages the expert's gradient history before matrix normalization.

We propose \mesh{}, a MoE-aware, hidden-momentum Sinkhorn optimizer. 
\mesh{} keeps the SAGE/Sinkhorn hybrid for vocabulary and non-expert matrices, but replaces stateless expert Sinkhorn with hidden-momentum expert Sinkhorn.  
Its expert first moment is maintained through the gradient-buffer lifecycle rather than as an optimizer-state tensor.  
We also evaluate \meshb{}, which adds a coarse block/neuron inverse-RMS multiplier.  
The main empirical message is deliberately conservative: MoE experts primarily need temporal smoothing before matrix normalization; block-level adaptivity can help, especially in seed-42 diagnostics, but is not the sole causal ingredient.

Our contributions are:
\begin{itemize}
\item We identify routed MoE experts as the failure point of direct SAGE/Sinkhorn optimization in nanowhale pretraining.
\item We introduce \mesh{}, a hidden-momentum Sinkhorn expert update that restores a first-moment signal without adding explicit expert first-moment optimizer state.
\item We evaluate \meshb{}, an optional block-preconditioned extension, and show that the unblocked and block-preconditioned hidden-momentum variants give similar multi-seed memory-quality trade-offs.
\item We provide negative ablations showing that second-moment-only, sign/momentum-only, confidence-only, scale-only, and fused-Sinkhorn interventions do not explain the gap.
\end{itemize}

%% file: files/related.tex
\subsection{Adaptive and memory-efficient optimization}
Adam and AdamW maintain first- and second-moment estimates for every parameter~\cite{kingma2015adam,loshchilov2019decoupled}.  
This robustness costs two state tensors per parameter.  Memory-efficient Adam variants reduce this cost in different ways: Adam-mini uses fewer learning-rate resources by exploiting block structure~\cite{zhang2024adammini}, and HMAdam/HMAdamW remove the explicit first-moment tensor by storing a momentum-like quantity in the gradient buffer~\cite{sadeghi2025hmadamw}.  
\mesh{} uses the latter idea only for MoE expert first moments and combines it with Sinkhorn-style matrix normalization.

\subsection{Matrix normalization and SAGE hybrids}
Sinkhorn and gradient multi-normalization optimizers compute matrix updates through row and column normalization~\cite{sinkhorn1964relationship,scetbon2025gradient}.  
Their appeal is that a large dense matrix can be updated without storing AdamW's two full state tensors.  
SAGE extends this strategy by treating embeddings as a separate difficult class~\cite{lee2026sage}.  
Our work identifies a second difficult class: routed MoE expert matrices.

\subsection{Mixture-of-Experts models}
Sparse MoE layers route tokens to a subset of experts, increasing model capacity without activating every parameter on every token~\cite{shazeer2017outrageously,lepikhin2021gshard,fedus2022switch}.  
This routing changes the optimizer problem.  
Expert gradients are conditional on router decisions and gate weights, so a routed expert matrix is not equivalent to a dense MLP matrix even when it has the same shape.  
\mesh{} is designed for this expert-specific gradient structure.

%% file: files/background.tex
\subsection{Sinkhorn updates for dense matrices}
Let $W\in\mathbb{R}^{m\times n}$ be a matrix parameter and $G_t=\nabla_W\mathcal{L}_t$ its gradient at step $t$.  A Sinkhorn-style matrix optimizer forms an update direction by repeatedly normalizing row and column norms,
\begin{equation}
    U_t = \mathrm{Sinkhorn}(G_t;K,\gamma),
\end{equation}
where $K$ is the number of normalization rounds and $\gamma$ is a scalar update scale.  
This update is attractive because it is stateless: it uses the current matrix gradient and stores no first or second moment for dense matrices.

\subsection{Why routed experts are not ordinary dense matrices}
A routed MoE expert weight is also a matrix, but its gradient has a different sampling structure.  
In a top-$k$ MoE layer, expert $e$ receives only tokens selected by the router:
\begin{equation}
    G_{e,t}=\sum_{i:e\in\mathrm{topk}(x_i,t)} a_{i,e,t}\,g_{i,e,t},
\end{equation}
where $a_{i,e,t}$ is the gate weight.  
Thus $G_{e,t}$ is a gradient under the conditional distribution $p(x\mid e\in\mathrm{topk}(x,t))$, not under the full mini-batch distribution.  
This conditional distribution changes as the router and hidden states change during training.

This matters because Sinkhorn is nonlinear.  
Dense training can often tolerate $\mathrm{Sinkhorn}(G_t)$ because $G_t$ is a reasonably stable estimate of a full-batch dense-matrix gradient.  
In routed experts, however, the sequence $G_{e,t}$ carries routing-induced temporal noise.  Applying Sinkhorn before smoothing computes $\mathrm{Sinkhorn}(G_{e,t})$ for each noisy expert sample.  
Momentum instead approximates applying Sinkhorn to a temporally averaged expert signal.  This distinction is the central diagnosis of the paper.

\subsection{Hypotheses we test}
We considered several explanations for the expert failure.  
Raw route-count sparsity could reduce effective batch size; gate-weight skew could produce effective sparsity even when every expert is selected; semantic routing could make expert gradients non-i.i.d.; Sinkhorn could be sensitive to expert tensor partitioning; or stateless Sinkhorn could simply lack temporal smoothing. 
Our experiments rule out the first four as complete explanations.  
The remaining supported hypothesis is that routed experts require a first-moment signal before matrix normalization.  
Coarse block/neuron preconditioning can improve this update, but the multi-seed results indicate that the first moment is the primary causal ingredient.

%% file: files/method.tex
\subsection{Problem: MoE expert gradients are not dense-matrix gradients}
Let $W_e$ be an expert matrix and $G_{e,t}=\nabla_{W_e}\mathcal{L}_t$ its gradient at step $t$.  
A dense MLP matrix receives gradient contributions from essentially all tokens in a minibatch.  A routed expert instead receives
\begin{equation}
    G_{e,t}=\sum_{i\in \mathcal{R}_{e,t}} a_{i,e,t}\, g_i(W_e),
\end{equation}
where $\mathcal{R}_{e,t}$ is the token subset selected for expert $e$ and $a_{i,e,t}$ is the gate weight.  
Both the subset and the weights change with the router and hidden states.  
Thus the expert gradient is a temporally varying conditional estimate.  
Applying stateless Sinkhorn directly to $G_{e,t}$ uses
\begin{equation}
    U_{e,t}=\mathrm{Sinkhorn}(G_{e,t}; K, \tau),
\end{equation}
which normalizes the instantaneous matrix without remembering whether the expert signal is persistent or a one-step routing fluctuation.

\begin{figure*}[t]
\centering
\fbox{\begin{minipage}{0.94\textwidth}
\small
\begin{tabular}{p{0.45\textwidth}p{0.45\textwidth}}
\toprule
\textbf{Stateless expert Sinkhorn} & \textbf{\mesh{}: average before normalization} \\
\midrule
$G_{e,t}\;\longrightarrow\;\mathrm{Sinkhorn}(G_{e,t})\;\longrightarrow\;$ update
&
$H_{e,t}=\beta_1H_{e,t-1}+G_{e,t}\;\longrightarrow\;\mathrm{Sinkhorn}(H_{e,t})\;\longrightarrow\;$ update
\\[0.5em]
Uses the instantaneous routed-expert gradient.  Routing fluctuations are normalized into updates.
&
Uses a temporally smoothed expert signal.  Routing fluctuations are averaged before matrix normalization.
\\
\bottomrule
\end{tabular}
\vspace{0.35em}
\centerline{\meshb{} additionally multiplies the Sinkhorn direction by a coarse mean-normalized block/neuron inverse-RMS factor.}
\end{minipage}}
\caption{Core MESH mechanism.  The failure is an ordering issue: normalizing each instantaneous routed-expert gradient is not equivalent to normalizing a temporally averaged expert signal.}
\label{fig:core}
\end{figure*}

\subsection{MESH: hidden momentum before Sinkhorn}
\method{} changes only the MoE expert role.  The primary algorithm, \mesh{}, maintains an unnormalized expert momentum accumulator
\begin{equation}
    H_{e,t}=\beta_1 H_{e,t-1}+G_{e,t}.
\end{equation}
This differs by a constant scale from the conventional Adam first moment.  
We use $H_{e,t}$ directly because Sinkhorn normalization and mean-normalized block multipliers are invariant to global scaling up to the explicit update scale. 
The expert matrix direction is then
\begin{equation}
    U_{e,t}=\mathrm{Sinkhorn}(H_{e,t};K,\tau),
\end{equation}
where $K$ is the number of alternating row/column normalization rounds and $\tau$ is the Sinkhorn update scale.

The hidden-momentum implementation stores $H_{e,t}$ outside the optimizer-state dictionary.  
After an optimizer step, the expert gradient buffer is stashed.  
Immediately before the next backward pass, the buffer is reattached as $p.grad\leftarrow \beta_1 H_{e,t}$.  
Backward accumulation adds the current gradient, yielding $H_{e,t+1}$.  
This prepare-before-backward lifecycle was necessary in Trainer/Accelerate/DDP-style loops, where a naive \texttt{zero\_grad} bridge can be bypassed by wrapped-model gradient clearing.  
Diagnostics in our final implementation report approximately one prepare call per optimizer step, buffer-shadow cosine near one, and relative error around $10^{-8}$.

\subsection{MESH-B: optional block/neuron scaling}
\mesh{} can be used as hidden-momentum Sinkhorn alone.  
We also evaluate a block-scaled variant, denoted \meshb{}, which applies a coarse inverse-RMS multiplier to the Sinkhorn direction.  
For expert $w_1,w_3$ matrices we use row statistics; for $w_2$ we use column statistics.  Let $R(H_{e,t})$ denote this block RMS.  
The default \meshb{} statistic is computed from the smoothed source $H_{e,t}$ rather than a separate raw-gradient stream:
\begin{equation}
    V_{e,t}=\beta_2 V_{e,t-1}+(1-\beta_2)R(H_{e,t})^2.
\end{equation}
We use $H$ for this statistic because the best seed-42 diagnostic used $V(H)$, and because it keeps the hidden-momentum implementation simple without storing an additional raw-gradient stream.  R
aw-gradient block statistics are a natural alternative, but we leave a systematic comparison to future work.

The multiplier is clipped and mean-normalized,
\begin{equation}
    Q_{e,t}=\mathrm{clip}\left(
    \frac{(V_{e,t}+\epsilon)^{-p/2}}{\mathrm{mean}((V_{e,t}+\epsilon)^{-p/2})}, q_{\min}, q_{\max}\right),
\end{equation}
and the expert update is
\begin{equation}
    W_{e,t+1}=W_{e,t}-\eta\, Q_{e,t}\odot U_{e,t}-\eta\lambda W_{e,t},
\end{equation}
with $Q_{e,t}$ broadcast over the appropriate block dimension.  
The block multiplier is intentionally coarse.  
It is not a full coordinate-wise second moment and is not the central causal claim of the paper.  
Our results suggest that temporal smoothing is the primary requirement, while block/neuron scaling can improve the frontier in some settings.

\begin{figure}[t]
\centering
\fbox{\begin{minipage}{0.88\linewidth}
\scriptsize
\textbf{MESH family role assignment.}\vspace{0.35em}

\setlength{\tabcolsep}{5pt}
\begin{tabular}{@{}p{0.38\linewidth}p{0.30\linewidth}p{0.22\linewidth}@{}}
\toprule
Parameter class & Optimizer role & Persistent state \\
\midrule
Embedding / LM head & SAGE & light state \\
Non-expert dense matrices & Sinkhorn & none \\
Routed/shared experts & \mesh{} & hidden buffer \\
Optional expert block scale & \meshb{} block RMS & small block state \\
Norms / biases & SAGE & light state \\
\bottomrule
\end{tabular}
\end{minipage}}
\caption{\method{} keeps the SAGE/Sinkhorn hybrid for non-expert parameters and replaces only the MoE expert update.  \mesh{} is hidden-momentum expert Sinkhorn; \meshb{} adds optional block preconditioning.  Full-state variants are diagnostic controls.}
\label{fig:roles}
\end{figure}

\subsection{Full-state diagnostic controls}
We use full-state variants only for diagnosis.  
\meshfull{} stores the same unnormalized expert momentum $H_{e,t}$ explicitly in optimizer state rather than in the gradient buffer.  
\meshbfull{} additionally stores the block statistic used by \meshb{}.  
Comparing hidden and full-state variants isolates the systems effect of hidden momentum from the optimizer effect of temporal smoothing.  
In our experiments, full-state variants are often stronger quality diagnostics, while \mesh{} and \meshb{} are the memory-efficient algorithms.

%% file: files/experiments.tex
\subsection{Setup}
We use the HuggingFace nanowhale training codebase, which implements a 110M-parameter DeepSeek-V4-style MoE language model~\cite{huggingface2025nanowhale}.
The model has eight layers, four routed experts and one shared expert per layer, hidden size 320, and an untied 129,280-token vocabulary. 
The vocabulary matrices are unusually large for this model size: input embedding plus LM head contain 82.75M of 110.37M trainable parameters.

Unless noted, runs train from scratch on FineWeb-Edu for 5,000 optimizer steps with sequence length 2048, per-device batch size 8, gradient accumulation 4, and eight data-parallel workers. 
This corresponds to an estimated 524,288 tokens per optimizer step and 2.62B model tokens per run.  
Evaluation uses 512 held-out examples.  
Because several differences are in the 0.01--0.02 loss range, we interpret small gaps as suggestive rather than conclusive.

SAGE/Sinkhorn and \mesh{} configurations follow the official SAGE-style hyperparameters unless stated otherwise: learning rate $2\times10^{-3}$, $\beta_1=0.9$, $\beta_2=0.99$, weight decay 0.01, cosine schedule, 10\% warmup, Sinkhorn scale $\tau=10$, and $K=5$ normalization rounds. 
The AdamW baseline uses the nanowhale-tuned configuration used in the project logs: learning rate $6\times10^{-4}$, $\beta_1=0.9$, $\beta_2=0.95$, weight decay 0.1, cosine schedule, and 3\% warmup.  
We report evaluation loss, mean token accuracy, persistent optimizer-state memory, and peak PyTorch CUDA allocation.

\begin{table*}[t]
\centering
\caption{Main multi-seed comparison.  These are the runs for which post-fix hidden-momentum results are available at seeds 100 and 512.  \mesh{} is hidden-momentum Sinkhorn without block preconditioning.  \meshb{} is \mesh{} with a block/neuron inverse-RMS multiplier.  State denotes persistent optimizer-state memory.}
\label{tab:main_multiseed}
\small
\begin{tabular}{lccccc}
\toprule
Optimizer & Seeds & Eval loss & Eval acc. & State (GB) & Peak CUDA (GB) \\
\midrule
AdamW & 100/512 & 3.5877 avg & 0.3515 avg & 0.883 & 4.375--4.376 \\
\meshb{} & 100/512 & 3.6365 avg & 0.3436 avg & 0.331 & 3.822--3.823 \\
\mesh{} & 100/512 & 3.6430 avg & 0.3430 avg & 0.331 & 3.823--3.824 \\
\bottomrule
\end{tabular}
\end{table*}

\begin{table}[t]
\centering
\caption{Seed-42 diagnostic comparison.  These single-seed ablations should not be directly averaged with Table~\ref{tab:main_multiseed}; they identify mechanisms.}
\label{tab:seed42_diag}
\small
\begin{tabular}{lcc}
\toprule
Optimizer & Eval loss & State (GB) \\
\midrule
AdamW & 3.6427 & 0.883 \\
SAGE/Sinkhorn exact & 3.8265 & 0.331 \\
Experts AdamW & 3.6158 & 0.528 \\
\meshfull{} & 3.6619 & 0.429 \\
\meshbfull{} & 3.6187 & 0.430 \\
\meshbfull{} with $V(H)$ & 3.6102 & 0.430 \\
Routed-only \meshbfull{} & 3.6353 & 0.410 \\
\bottomrule
\end{tabular}
\end{table}

\subsection{Main results}
Table~\ref{tab:main_multiseed} shows that \meshb{} achieves a strong memory-quality trade-off but does not outperform AdamW in the additional seeds.
Relative to AdamW, \meshb{} reduces optimizer-state memory by 62.5\% and peak PyTorch CUDA allocation by about 12.6\%, but trails AdamW by about 0.049 evaluation loss.  
\mesh{} without the block multiplier is close to \meshb{}, suggesting that hidden temporal smoothing is the dominant ingredient and that block preconditioning is an incremental improvement in these seeds.

Table~\ref{tab:seed42_diag} separates diagnostic seed-42 ablations from the multi-seed table. 
In this seed, full-state \meshb{} variants recover AdamW-like quality with about half of AdamW's optimizer-state memory.  We use these runs as mechanistic evidence rather than as final multi-seed performance claims.

\subsection{Where the SAGE/Sinkhorn hybrid fails}
\begin{table}[t]
\centering
\caption{Expert role ablation, seed 42.}
\label{tab:expertroles}
\small
\begin{tabular}{lcc}
\toprule
Expert treatment & Eval loss & State (GB) \\
\midrule
All experts Sinkhorn & 3.8265 & 0.331 \\
Routed experts AdamW & 3.6297 & 0.488 \\
Shared expert AdamW & 3.6895 & 0.370 \\
Routed+shared AdamW & 3.6158 & 0.528 \\
Routed+shared SAGE & 3.8912 & 0.430 \\
\bottomrule
\end{tabular}
\end{table}

Table~\ref{tab:expertroles} shows that routed experts are the dominant failure point. 
Converting only routed experts to AdamW recovers most of the gap, while converting only the shared expert helps less.  
Applying SAGE to all experts worsens training, indicating that the embedding-oriented SAGE update is not a suitable expert replacement.

\subsection{What the expert update needs}
\begin{table}[t]
\centering
\caption{Expert update decomposition, seed 42.  These results motivate \mesh{} but do not imply that every component is universally necessary.}
\label{tab:decomposition}
\small
\begin{tabular}{lcc}
\toprule
Expert update & Eval loss & State (GB) \\
\midrule
Instant Sinkhorn & 3.8265 & 0.331 \\
V-only AdamW & 3.8457 & 0.429 \\
Lion / sign-momentum & 4.0046 & 0.429 \\
Factored-v only & 4.2308 & 0.331 \\
\meshfull{} & 3.6619 & 0.429 \\
\meshfull{}, $\beta_1=0.7$ & 3.6438 & 0.429 \\
\meshbfull{} & 3.6187 & 0.430 \\
\meshbfull{}, $V(H)$ & 3.6102 & 0.430 \\
\bottomrule
\end{tabular}
\end{table}

Table~\ref{tab:decomposition} shows that adding a temporal first moment to expert Sinkhorn is the major improvement. 
V-only, sign-only, and factored-v controls are poor.  
Block/neuron preconditioning can improve the best single-seed results, but the closeness of \mesh{} and \meshb{} in Table~\ref{tab:main_multiseed} prevents a stronger claim that block preconditioning is universally required.

\subsection{Memory accounting}
AdamW stores two full moment states for all 110.37M parameters, producing 0.883GB optimizer state.  
Direct SAGE/Sinkhorn stores state mainly for the 82.75M vocabulary/SAGE parameters, producing 0.331GB. 
\mesh{} keeps this same optimizer-state size: the expert first moment is carried by the gradient-buffer lifecycle and counted separately as a 0.098GB hidden buffer, not as optimizer state.  
Although the hidden buffer is excluded from optimizer-state accounting because it lives in the gradient-buffer lifecycle, it is still part of the persistent training memory and is reflected in CUDA peak measurements.
\meshb{} adds only small block statistics at this model size, so its reported optimizer-state memory remains 0.331GB at three-decimal precision.
\meshbfull{} stores the expert first moment explicitly, increasing state to about 0.430GB.  Peak CUDA allocation decreases less than optimizer-state memory because parameters, activations, communication buffers, gradient buffers, temporary update tensors, and allocator effects remain.

%% file: files/analysis.tex
\subsection{Why routed experts needed momentum}
The routed expert failure is best understood as an ordering problem.  
Stateless Sinkhorn applies $\mathrm{Sinkhorn}(G_t)$ at every step.  
\mesh{} applies $\mathrm{Sinkhorn}(H_t)$, where $H_t$ averages the expert gradient sequence over time.  Because Sinkhorn is nonlinear, $\mathrm{Sinkhorn}(\mathbb{E}G)$ is not equivalent to $\mathbb{E}[\mathrm{Sinkhorn}(G)]$.  
For routed experts, this difference is large: averaging first filters routing-induced temporal noise, and only then applies matrix normalization.  
This explains why momentum Sinkhorn improves substantially over instantaneous Sinkhorn, even before adding a block multiplier.

The need for momentum is specific to the routed path.  
A shared expert is active for every token and behaves more like a dense MLP path.  In contrast, a routed expert sees a router-conditioned token subset with learned gate weights.  
Its gradient is therefore a sequence of conditional samples rather than a sequence of dense minibatch estimates.  
Momentum tracks the expert's persistent direction across these samples.

\subsection{What block preconditioning contributes}
Block/neuron inverse-RMS scaling improves some single-seed settings.  
In seed 42, \meshbfull{} reaches 3.6187 eval loss, and the $V(H)$ variant reaches 3.6102.  However, in seeds 100 and 512, hidden-momentum Sinkhorn without block preconditioning is close to the block-preconditioned \meshb{} result.  
We therefore treat block preconditioning as a useful coarse adaptive scale, not as a conclusively necessary component.  
The safest conclusion is that MoE expert optimization primarily requires temporal smoothing; block/neuron adaptivity can improve the frontier and deserves further study.

\subsection{Residual gap to AdamW}
The strongest AdamW seeds remain better than \mesh{} or \meshb{} in the current experiment.  
This residual gap may not be caused by expert optimization alone.  
The nanowhale configuration is vocabulary-heavy: input embedding plus untied LM head account for approximately 75\% of trainable parameters.  
\mesh{} leaves those matrices on SAGE, while AdamW adapts them with full first and second moments.  Non-expert dense matrices also remain on stateless Sinkhorn. 
Future role ablations should therefore keep \mesh{} on experts while assigning only vocabulary matrices or only non-expert dense matrices to AdamW.

\subsection{Limitations}
The study is intentionally narrow.  
It uses one 110M MoE architecture, one streaming corpus, 5,000-step runs, and primarily 512-example evaluations.  
Several gaps are small enough that larger evaluation sets and more seeds are needed.  
We also do not claim that \mesh{} outperforms AdamW.  Rather, \mesh{} provides a substantially lower-state point on the memory-quality frontier and identifies temporal smoothing before matrix normalization as a key missing ingredient for MoE expert Sinkhorn optimization.

%% file: files/conclusion.tex
We presented \mesh{}, a hidden-momentum Sinkhorn optimizer for MoE expert matrices, and \meshb{}, an optional block-preconditioned variant.  
The study shows that direct SAGE/Sinkhorn optimization fails on routed MoE experts because the expert gradient sequence is conditional and temporally noisy.  
The missing ingredient is primarily temporal smoothing before matrix normalization.  Hidden momentum supplies this signal without storing an explicit expert first-moment tensor in optimizer state.  
The resulting algorithms reduce optimizer-state memory substantially relative to AdamW while recovering much of the quality lost by stateless Sinkhorn.  
The remaining gap to AdamW likely involves non-expert role assignment, especially vocabulary matrices, and motivates future work on memory-efficient updates for the entire MoE parameter taxonomy.

%% file: files/appendix_results.tex
\subsection{Full result tables}
Table~\ref{tab:app_multiseed} contains the multi-seed comparison used in the main text.  
Table~\ref{tab:app_expert} collects single-seed expert-update ablations.  
Table~\ref{tab:app_routing} collects routing controls.  
Table~\ref{tab:app_negative} lists negative or partial causal interventions.  
AdamW baselines use learning rate $6\times10^{-4}$, $\beta_1=0.9$, $\beta_2=0.95$, weight decay 0.1, cosine decay, and 3\% warmup.  
SAGE/MESH-style runs use learning rate $2\times10^{-3}$, $\beta_1=0.9$, $\beta_2=0.99$, weight decay 0.01, cosine decay, 10\% warmup, Sinkhorn scale 10, and five Sinkhorn rounds unless otherwise stated.

\begin{table*}[h]
\centering
\smalltab
\setlength{\tabcolsep}{4pt}
\begin{tabular}{lccccc}
\toprule
Optimizer & Seed & Eval loss & Eval acc. & State GB & CUDA peak GB \\
\midrule
AdamW & 42 & 3.6427 & 0.3464 & 0.883 & 4.426 \\
AdamW & 100 & 3.5848 & 0.3520 & 0.883 & 4.376 \\
AdamW & 512 & 3.5906 & 0.3510 & 0.883 & 4.375 \\
\meshb{} & 100 & 3.6335 & 0.3450 & 0.331 & 3.822 \\
\meshb{} & 512 & 3.6395 & 0.3422 & 0.331 & 3.823 \\
\mesh{} & 100 & 3.6364 & 0.3433 & 0.331 & 3.824 \\
\mesh{} & 512 & 3.6495 & 0.3426 & 0.331 & 3.823 \\
\bottomrule
\end{tabular}
\caption{Post-fix hidden-momentum multi-seed comparison.  \mesh{} denotes hidden-momentum expert Sinkhorn without block preconditioning; \meshb{} denotes the optional block-preconditioned variant.}
\label{tab:app_multiseed}
\end{table*}

\begin{table*}[h]
\centering
\smalltab
\setlength{\tabcolsep}{3.5pt}
\begin{tabular}{lccp{6.0cm}}
\toprule
Expert update & Eval loss & State GB & Interpretation \\
\midrule
SAGE/Sinkhorn exact & 3.8265 & 0.331 & Stateless expert Sinkhorn fails. \\
Routed expert AdamW & 3.6297 & 0.488 & Routed experts explain most of the expert gap. \\
Shared expert AdamW & 3.6895 & 0.370 & Shared expert matters less. \\
Routed+shared AdamW & 3.6158 & 0.528 & Strong expert-only upper bound. \\
Experts SAGE & 3.8912 & 0.430 & Embedding-oriented SAGE is unsuitable for experts. \\
V-only AdamW & 3.8457 & 0.429 & Second moment without first moment is insufficient. \\
Lion / sign-momentum & 4.0046 & 0.429 & Sign-only update is insufficient. \\
Factored-v only & 4.2308 & 0.331 & Row/column factored scale alone is too weak. \\
Block AdamW neuron & 3.6268 & 0.430 & Block/neuron scale with momentum is strong. \\
\meshfull{} & 3.6619 & 0.429 & Temporal smoothing explains much of the recovery. \\
\meshfull{} $\beta_1=0.7$ & 3.6438 & 0.429 & Smoothing strength matters. \\
\meshbfull{} $p=0.5$ & 3.6187 & 0.430 & Strong full-state diagnostic. \\
\meshbfull{} $V(H)$ & 3.6102 & 0.430 & Best seed-42 diagnostic variant. \\
\bottomrule
\end{tabular}
\caption{Expert optimizer decomposition, seed 42 unless marked otherwise.  Full-state variants are diagnostic controls; the proposed memory-efficient algorithms are hidden-momentum variants in Table~\ref{tab:app_multiseed}.}
\label{tab:app_expert}
\end{table*}

\begin{table*}[h]
\centering
\smalltab
\setlength{\tabcolsep}{4pt}
\begin{tabular}{llccc}
\toprule
Routing mode & Optimizer & Eval loss & State GB & Peak GB \\
\midrule
Top-1 learned & AdamW & 3.6777 & 0.883 & 4.198 \\
Top-1 learned & SAGE/Sinkhorn & 3.8710 & 0.331 & 3.644 \\
Top-2 learned & AdamW & 3.6427 & 0.883 & 4.426 \\
Top-2 learned & SAGE/Sinkhorn & 3.8265 & 0.331 & 3.877 \\
Top-3 learned & AdamW & 3.5816 & 0.883 & 4.553 \\
Top-3 learned & SAGE/Sinkhorn & 3.7590 & 0.331 & 4.000 \\
All-top-k learned & AdamW & 3.6281 & 0.883 & 4.777 \\
All-top-k learned & SAGE/Sinkhorn & 3.8004 & 0.331 & 4.225 \\
All-uniform & AdamW & 3.6337 & 0.883 & 4.639 \\
All-uniform & SAGE/Sinkhorn & 3.7563 & 0.331 & 4.036 \\
Random balanced top-2 & SAGE/Sinkhorn & 3.8050 & 0.331 & -- \\
\bottomrule
\end{tabular}
\caption{Routing and dense-like controls.  Top-k and routing interventions do not remove the SAGE/Sinkhorn gap.}
\label{tab:app_routing}
\end{table*}

\begin{table*}[h]
\centering
\smalltab
\setlength{\tabcolsep}{4pt}
\begin{tabular}{lccp{7.0cm}}
\toprule
Intervention & Eval loss & State GB & Takeaway \\
\midrule
Expert Sinkhorn scale 5 & 3.8212 & 0.331 & Not just an update-scale issue. \\
Scale-preserved Sinkhorn & 3.7632 & 0.331 & Re-inserting raw gradient scale is insufficient. \\
Confidence damping & 3.7605 & 0.331 & Route confidence is secondary. \\
Confidence amplification & diverged & 0.331 & Direct route-confidence amplification is unsafe. \\
Fused routed Sinkhorn & 3.7662 & 0.331 & Expert tensor partitioning is not primary. \\
Fused routed+shared, all-uniform & 3.7180 & 0.331 & Fusion helps slightly but is not enough. \\
\bottomrule
\end{tabular}
\caption{Negative and partial interventions.}
\label{tab:app_negative}
\end{table*}

%% file: files/appendix_algorithm.tex
\subsection{Sinkhorn normalization}
\begin{framed}
\noindent\textbf{Algorithm 1: SinkhornNormalize}$(G,K,\epsilon)$
\begin{algorithmic}[1]
\STATE $X\leftarrow G$
\FOR{$i=1$ to $K$}
    \STATE $X\leftarrow X/(\|X_{r:}\|_2+\epsilon)$ \COMMENT{row normalization}
    \STATE $X\leftarrow X/(\|X_{:c}\|_2+\epsilon)$ \COMMENT{column normalization}
\ENDFOR
\RETURN $X$
\end{algorithmic}
\end{framed}

\subsection{MESH expert step}
\begin{framed}
\noindent\textbf{Algorithm 2: MESH expert update}
\begin{algorithmic}[1]
\REQUIRE expert parameter $W$, hidden stash $H$, learning rate $\eta$, weight decay $\lambda$, Sinkhorn scale $\gamma$
\STATE \textbf{Before backward:} if $H$ exists, set $\mathrm{grad}(W)\leftarrow \beta_1 H$; otherwise set $\mathrm{grad}(W)\leftarrow 0$.
\STATE Run backward.  The gradient slot now contains $H_t=\beta_1H_{t-1}+G_t$.
\STATE $D_t\leftarrow \mathrm{SinkhornNormalize}(H_t,K,\epsilon)$.
\STATE $W\leftarrow W-\eta\gamma D_t-\eta\lambda W$.
\STATE Stash $H\leftarrow \mathrm{detach}(H_t)$ outside the optimizer-state dictionary.
\STATE Clear the gradient slot normally.
\end{algorithmic}
\end{framed}

\subsection{Optional MESH-B block-preconditioned expert step}
\begin{framed}
\noindent\textbf{Algorithm 3: MESH-B expert update}
\begin{algorithmic}[1]
\REQUIRE expert parameter $W$, hidden accumulator $H_t$, block state $V$, power $p$
\STATE $D_t\leftarrow \mathrm{SinkhornNormalize}(H_t,K,\epsilon)$.
\STATE $B_t\leftarrow \rho(H_t)$, where $\rho$ is row RMS for $W_1/W_3$ and column RMS for $W_2$.
\STATE $V\leftarrow \beta_2V+(1-\beta_2)B_t^2$.
\STATE $P\leftarrow \mathrm{clip}\left(\frac{(V+\epsilon)^{-p/2}}{\mathrm{mean}((V+\epsilon)^{-p/2})},c_{\min},c_{\max}\right)$.
\STATE $W\leftarrow W-\eta\gamma(P\odot D_t)-\eta\lambda W$.
\end{algorithmic}
\end{framed}
We use $H_t$ rather than a separate raw-gradient stream for the default block statistic.  
This matches the strongest seed-42 diagnostic, denoted $V(H)$ in the tables, and avoids storing an additional full raw-gradient stream in the hidden-momentum implementation.

\subsection{Full-state diagnostic variants}
The full-state diagnostics, \meshfull{} and \meshbfull{}, store $H_t$ explicitly in optimizer state rather than through the gradient buffer.  
\meshfull{} is otherwise identical to Algorithm~2, while \meshbfull{} is otherwise identical to Algorithm~3.  
We use these variants to separate the value of temporal smoothing from the engineering question of whether hidden momentum can faithfully reproduce the explicit first moment.

\subsection{Hidden momentum scale convention}
The hidden buffer obeys $H_t=\beta_1H_{t-1}+G_t$, while a conventional Adam-style first moment uses $M_t=\beta_1M_{t-1}+(1-\beta_1)G_t$.  
Thus $H_t$ is an unnormalized first-moment accumulator.  
Since Sinkhorn normalization and mean-normalized block multipliers are invariant to global scaling up to the explicit update scale, we use $H_t$ directly.

\subsection{Trainer lifecycle}
A naive \texttt{optimizer.zero\_grad()} override was insufficient in our training stack because the Trainer, Accelerate, or distributed wrapper can clear gradients through model-level paths.  
The final implementation uses a prepare-before-backward hook.  
The optimizer stashes hidden expert buffers after each step; immediately before backward, the trainer reattaches the decayed stash to the gradient slot.  
Diagnostics on full runs report approximately one prepare call per optimizer step, buffer-shadow cosine near one, and relative error around $10^{-8}$.

\subsection{Role counts in the main MESH configurations}
In the main \mesh{} configuration, 85 tensors with 82.747M parameters use SAGE, 65 non-expert dense tensors with 3.046M parameters use plain Sinkhorn, and 120 expert tensors with 24.576M parameters use hidden-momentum expert Sinkhorn.  
In \meshb{}, the same 120 expert tensors additionally use the block/neuron multiplier.  
The hidden expert buffer is estimated at 0.098GB.  
\meshb{} has the same expert coverage and adds small block statistics.  
The optimizer-state footprint is 0.331GB at three-decimal precision, dominated by the vocabulary/SAGE states.

%% file: Template.bbl
{\small
\begin{thebibliography}{10}
\bibitem{kingma2015adam}
D.~P. Kingma and J.~Ba, ``Adam: A method for stochastic optimization,'' in \emph{International Conference on Learning Representations}, 2015.

\bibitem{loshchilov2019decoupled}
I.~Loshchilov and F.~Hutter, ``Decoupled weight decay regularization,'' in \emph{International Conference on Learning Representations}, 2019.

\bibitem{sinkhorn1964relationship}
R.~Sinkhorn, ``A relationship between arbitrary positive matrices and doubly stochastic matrices,'' \emph{Annals of Mathematical Statistics}, vol.~35, no.~2, pp.~876--879, 1964.

\bibitem{scetbon2025gradient}
M.~Scetbon, C.~Ma, W.~Gong, and E.~Meeds, ``Gradient multi-normalization for efficient LLM training,'' \emph{OpenReview}, 2025. Available: \url{https://openreview.net/forum?id=oanhUGY6un}

\bibitem{lee2026sage}
W.~Lee and H.-T. Kim, ``SAGE: Sign-adaptive gradient for memory-efficient LLM optimization,'' \emph{arXiv:2604.07663}, 2026.

\bibitem{sadeghi2025hmadamw}
M.~A. Sadeghi, ``Eliminating the first moment state in Adam optimizer,'' \emph{OpenReview}, 2025. Available: \url{https://openreview.net/forum?id=xRxh48OAAM}

\bibitem{zhang2024adammini}
Y.~Zhang \emph{et al.}, ``Adam-mini: Use fewer learning rates to gain more,'' \emph{arXiv:2406.16793}, 2024.

\bibitem{shazeer2017outrageously}
N.~Shazeer \emph{et al.}, ``Outrageously large neural networks: The sparsely-gated mixture-of-experts layer,'' in \emph{International Conference on Learning Representations}, 2017.

\bibitem{lepikhin2021gshard}
D.~Lepikhin \emph{et al.}, ``GShard: Scaling giant models with conditional computation and automatic sharding,'' in \emph{International Conference on Learning Representations}, 2021.

\bibitem{fedus2022switch}
W.~Fedus, B.~Zoph, and N.~Shazeer, ``Switch Transformers: Scaling to trillion parameter models with simple and efficient sparsity,'' \emph{Journal of Machine Learning Research}, 2022.

\bibitem{huggingface2025nanowhale}
Hugging Face, ``nanowhale: A 110M-parameter DeepSeek-style language model,'' GitHub repository, 2025. Available: \url{https://github.com/huggingface/nanowhale}
\end{thebibliography}}
